\documentclass[10pt,twocolumn,letterpaper]{article}

\usepackage{3dv}
\usepackage{times}
\usepackage{epsfig}
\usepackage{graphicx}
\usepackage{amsmath}
\usepackage{amssymb}

\usepackage{soul}
\usepackage{overpic}
\usepackage{makecell}
\usepackage{scalerel}

\usepackage[pagebackref=true,breaklinks=true,colorlinks,bookmarks=false]{hyperref}

\threedvfinalcopy 

\def\threedvPaperID{86} 

\ifthreedvfinal\fi
\begin{document}

\title{Dance In the Wild: Monocular Human Animation with Neural Dynamic Appearance Synthesis}

\author{
Tuanfeng Y. Wang \qquad Duygu Ceylan \qquad Krishna Kumar Singh  \\
Adobe Research\\
{\tt\small \{yangtwan,ceylan,krishsin\}@adobe.com}
\and Niloy J. Mitra \\
Adobe Research, University College London\\
{\tt\small nimitra@adobe.com}}
\maketitle
\thispagestyle{empty}

\begin{abstract}
Synthesizing dynamic appearances of humans in motion plays a central role in applications such as AR/VR and video editing. While many recent methods have been proposed to tackle this problem, handling loose garments with complex textures and high dynamic motion still remains challenging. 
In this paper, we propose a video based appearance synthesis method that tackles such challenges and demonstrates high quality results for in-the-wild videos that have not been shown before. Specifically, we adopt a StyleGAN based architecture to the task of person specific video based motion retargeting. We introduce a novel motion signature that is used to modulate the generator weights to capture dynamic appearance changes as well as regularizing the single frame based pose estimates to improve temporal coherency. We evaluate our method on a set of challenging videos and show that our approach achieves state-of-the-art performance both qualitatively and quantitatively. 

\end{abstract}

\section{Introduction}
Generating plausible video based animations of an actor has several applications in AR/VR and video editing. A common approach to tackle this generation task is to transfer or retarget a motion sequence extracted from a \emph{source} video to the \emph{target} actor. Such approaches often depend on learning an actor specific appearance model from a training performance of the actor which is later utilized to synthesize the actor in novel poses. 

While many recent approaches that leverage advances in machine learning have been proposed to the video-based performance retargeting problem, extending them to \emph{in-the-wild} scenarios still remains challenging. First of all, many methods make simplifying assumptions about the appearance of the actor such as wearing tight clothing. Some recent work that extend to actors wearing loose garments~\cite{Kappel_2021_CVPR}, on the other hand, requires very long training videos of the actor which limit the applicability of the approach. Furthermore, such methods are often limited in terms of the complexity of the motions they can handle. We observe that with loose garments and complex motion sequences, state-of-the-art pose estimation methods~\cite{openpose,densepose} often suffer from wrong predictions, missing parts, and temporal noise; making it particularly challenging to learn a consistent appearance model for the actor which can capture motion dependent appearance changes.

\begin{figure}[!t]
  \includegraphics[width=\columnwidth]{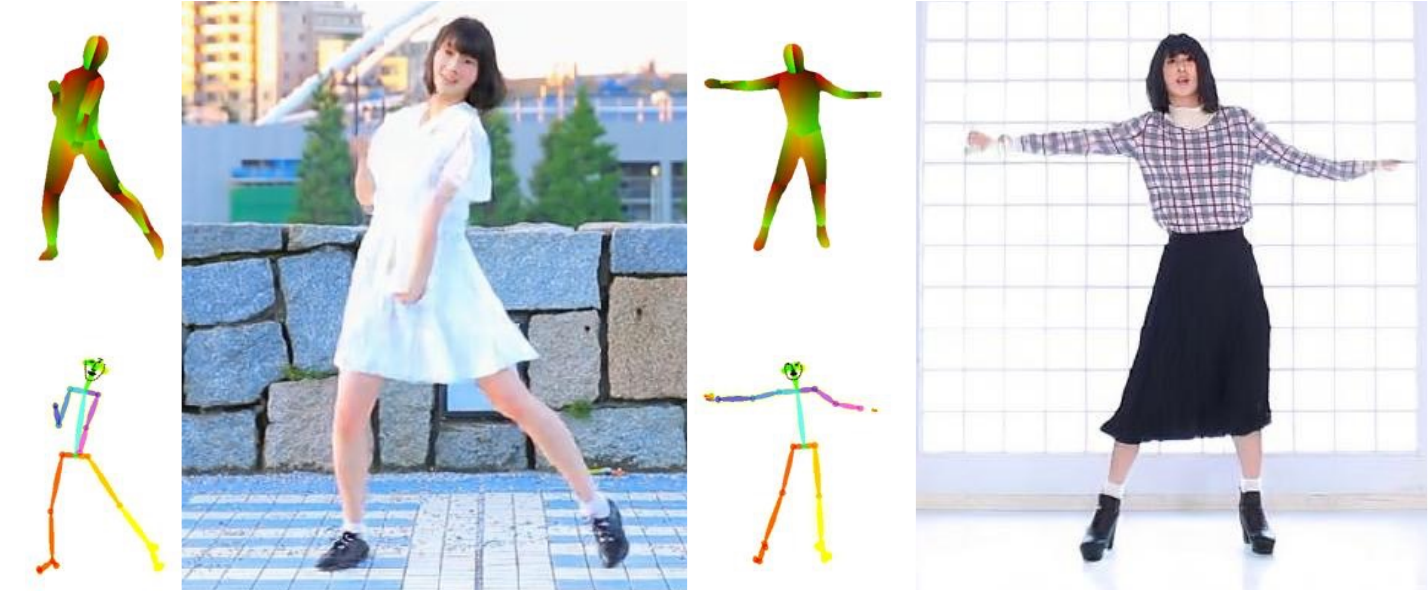}
  \caption{We present a method to synthesize the dynamic appearance of an actor given a target motion sequence. Our approach works well for actors wearing loose garments with complex textures and performing complex motions.}
  \label{fig:teaser}
  \vspace*{-5mm}
\end{figure}

In this work, we propose a novel approach to learn the dynamic appearance of an actor, potentially wearing loose garments (which may be far away from the body and may have significant dynamics under motion), and synthesize unseen complex motion sequences (see Fig.~\ref{fig:teaser}). Our approach adopts the state-of-the-art photo-realistic generative models for human portraits and bodies, such as StyleGAN~\cite{Karras19,Karras20} and StylePoseGAN~\cite{styleposegan}, to the task of person-specific appearance modeling and motion retargeting. In order to disentangle the pose and appearance of the actor, we utilize off-the-shelf keypoint~\cite{openpose} or dense correspondence~\cite{densepose} estimations to represent the target pose and provide as spatial features to the StyleGAN architecture. The core of our approach is to learn an effective motion representation that is used to demodulate the weights of the generator. This modulation not only helps to capture the appearance of loose garments that heavily depend on the underlying body motion but also captures plausible motion specific appearance changes (see Fig.~\ref{fig:disentangle}). Since keypoint or dense correspondence estimation methods operate on single images, they often result in pose representations that are not perfect and temporally coherent. We show that we can also use the motion features to refine the per-frame pose features and provide significant improvements in terms of temporal coherency as well.

We evaluate our method on several challenging motion sequences as well as target actors with loose garments consisting of complex texture patterns. We show that our results can synthesize plausible garment deformations while also maintaining high quality visual results. We also perform extensive qualitative and quantitative comparisons with previous methods and demonstrate state of the art results. In summary, our contributions are as follows: 
\begin{itemize}
    \item We adopt the StyleGAN architecture to the task of actor specific video based motion retargeting. We demonstrate results that capture the dynamic appearance of loose garments under complex motion sequences that have not been shown before.
    \item We introduce an explicit motion representation that is used for demodulation of the generator weights. Such motion features are used to both capture motion specific appearance changes as well as generating temporally coherent results.
\end{itemize}

%
%
%
\section{Related Work}
Rendering of humans in different poses, from different viewpoints has attracted a lot of attention both from the graphics and vision communities. Below, we review the previous work most related to our method.

\textbf{Rendering based approaches.} Given very high quality geometry and appearance representations, a traditional approach to controllable synthesis of humans is to employ classical rendering techniques. Hence, several works have focused on reconstructing such geometry and appearance representations from multi-view captures of humans~\cite{Volino:BMVC:2014,Casas2014,Collet:2015}. With the recent success of deep learning, more recent methods have replaced certain components of such a pipeline with neural networks. For example, given a person specific 3D template of an actor, some methods learn dynamic appearances of the actor with neural networks~\cite{Liu2019Neural,liu2020NeuralHumanRendering,habermann2021,zhang2021dynamic}. Although providing explicit control, the 3D template requirement limits the application of such methods.
Another recent thread of work has explored the use of \emph{neural textures}~\cite{thies2019deferred,Grigorev_2021_CVPR} and \emph{neural rendering}~\cite{Tewari2020} to synthesize humans under different poses and viewpoints~\cite{Shysheya_2019_CVPR,Sarkar2020,raj2021anr,peng2021neural}. Most of these approaches, however, assume humans wear tight clothing and cannot handle complex motion sequences as our method.

\textbf{Image generation methods.} Image generation methods have gained significant popularity in the human reposing problem, i.e., generating a new image of a person given a source image and a target pose. Approaches that synthesize images from target poses represented as keypoints~\cite{ma2017pose,PumarolaASM18,Siarohin_2018_CVPR,zhu2019progressive}, dense uv coordinates~\cite{neverova2018,grigorev2019coordinate,SMPLpix:WACV:2020}, and heatmaps~\cite{ma2017disentangled} have been presented. Some methods have explored the use of warping techniques along with GANs~\cite{lwb2019} to improve the quality of the results. In the context of synthesizing human performance videos, while some approaches have explored temporal cues as training losses~\cite{chan2019dance,Aberman2019}, others have also utilized the temporal information at inference time~\cite{wang2018video,RecycleGAN}. In both single image and video-based generation methods, one problem that has been relatively less explored is to handle actors with loose clothing. Very recently, Kappel et al.~\cite{Kappel_2021_CVPR} have presented a recurrent network architecture that performs video-based animation for actors wearing loose garments. Starting from a set of 2D keypoints, they predict semantic part labels, a structure map that encodes garment wrinkles and texture patterns, and the final appearance of the actor. While showing impressive results, our experiments show that the method requires relatively long training sequences and is limited in terms of the complexity of the motions it can handle.

In the context of face generation, StyleGAN~\cite{Karras19,Karras20} has shown impressive results and many follow up work has focused on providing control in the latent space of the generator~\cite{tewari2020stylerig}. Compared to faces, however, controllable synthesis of full human bodies is more challenging. Very recently, StylePoseGAN~\cite{styleposegan}, VOGUE~\cite{lewis2020vogue}, and TryOnGAN~\cite{lewis2021tryongan} present earlier examples of adopting the StyleGAN architecture to the tasks of reposing and virtual try on. While our work is inspired by such approaches, we focus on the different and challenging task of dynamic motion synthesis.

\section{Methodology}

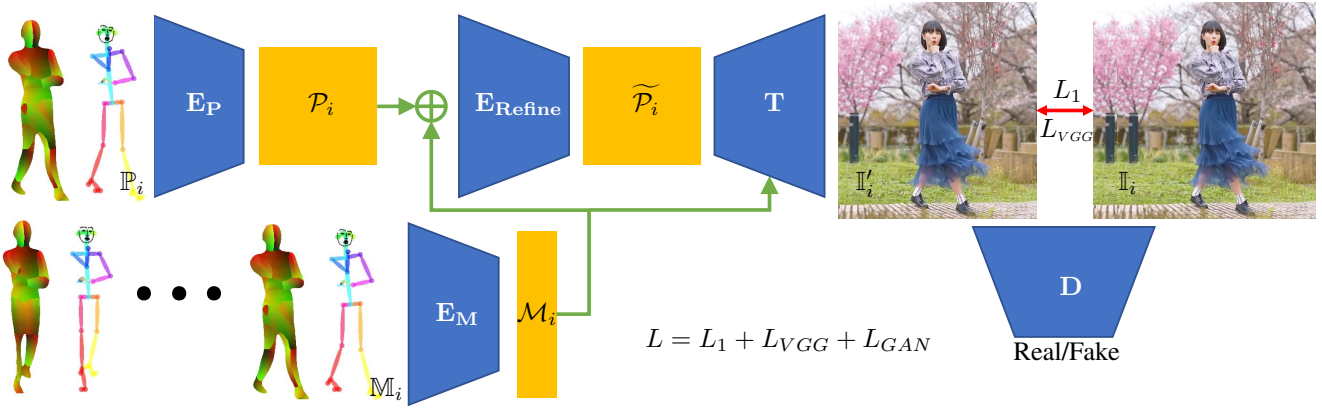
\begin{figure*}[!ht]
  \begin{overpic}[width=\textwidth]{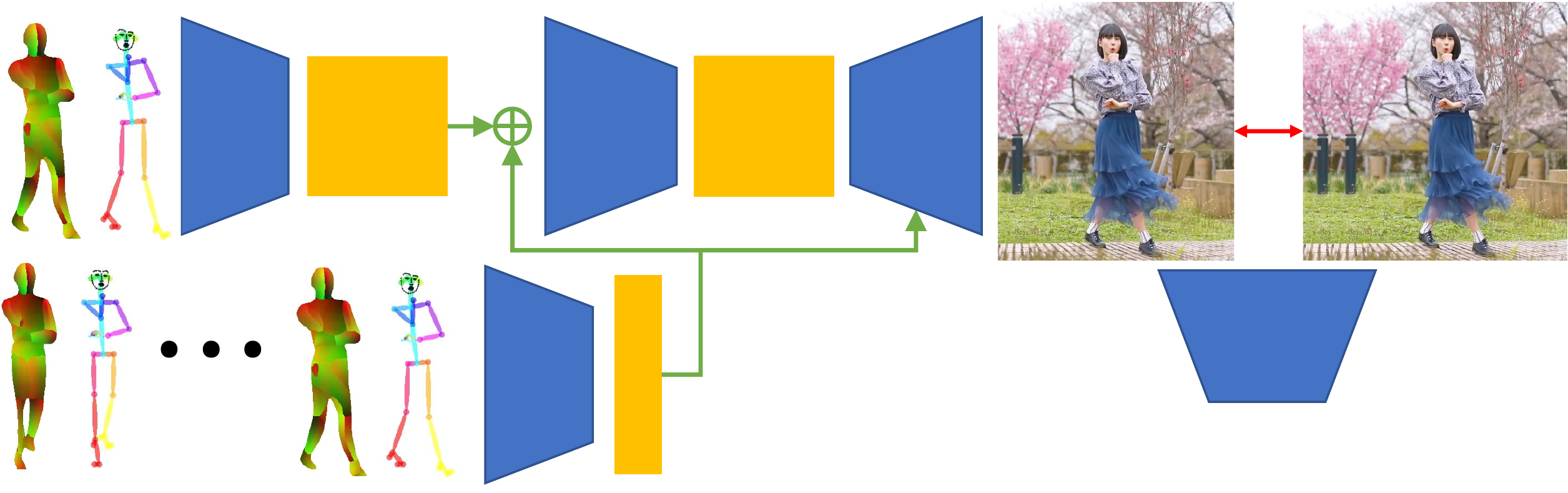}
    \put(9,17){$\mathbb{P}_i$}
    \put(28,1.3){$\mathbb{M}_i$}
    \put(14,23){\color{white}$\mathbf{E}_\mathbf{P}$}
    \put(23.3,23){$\mathcal{P}_i$}
    \put(33,7){\color{white}$\mathbf{E}_\mathbf{M}$}
    \put(39.2,7){$\mathcal{M}_i$}
    \put(36,23){\color{white}$\mathbf{E}_\mathbf{Refine}$}
    \put(48,23){$\widetilde{\mathcal{P}_i}$}
    \put(58,23){\color{white}$\mathbf{T}$}
    \put(65,17){$\mathbb{I}'_i$}
    \put(85,17){$\mathbb{I}_i$}
    \put(80.5,9){\color{white}$\mathbf{D}$}
    \put(77,4){Real/Fake}
    \put(49,5){$L = L_1 + L_{VGG} + L_{GAN}$}
    \put(80,24){$L_1$}
    \put(79,21){$L_{\scaleto{VGG\mathstrut}{5pt}}$}
    
  \end{overpic}
  
  \caption{Pipeline overview. Our network takes the 2D body dense UV and keypoints as input and learns a pose feature for each frame. By concatenating the pose inputs for the past few frames, we also learn motion features. The learned motion features are used to refine the pose features to improve the temporal coherency. We synthesize the final motion-aware dynamic appearance of the character using a StyleGAN based generator conditioned on the refined pose features and modulated by the motion features.}

  \label{fig:arch}
\end{figure*}

\subsection{Overview}
\begin{figure}[!h]
  \includegraphics[width=\columnwidth]{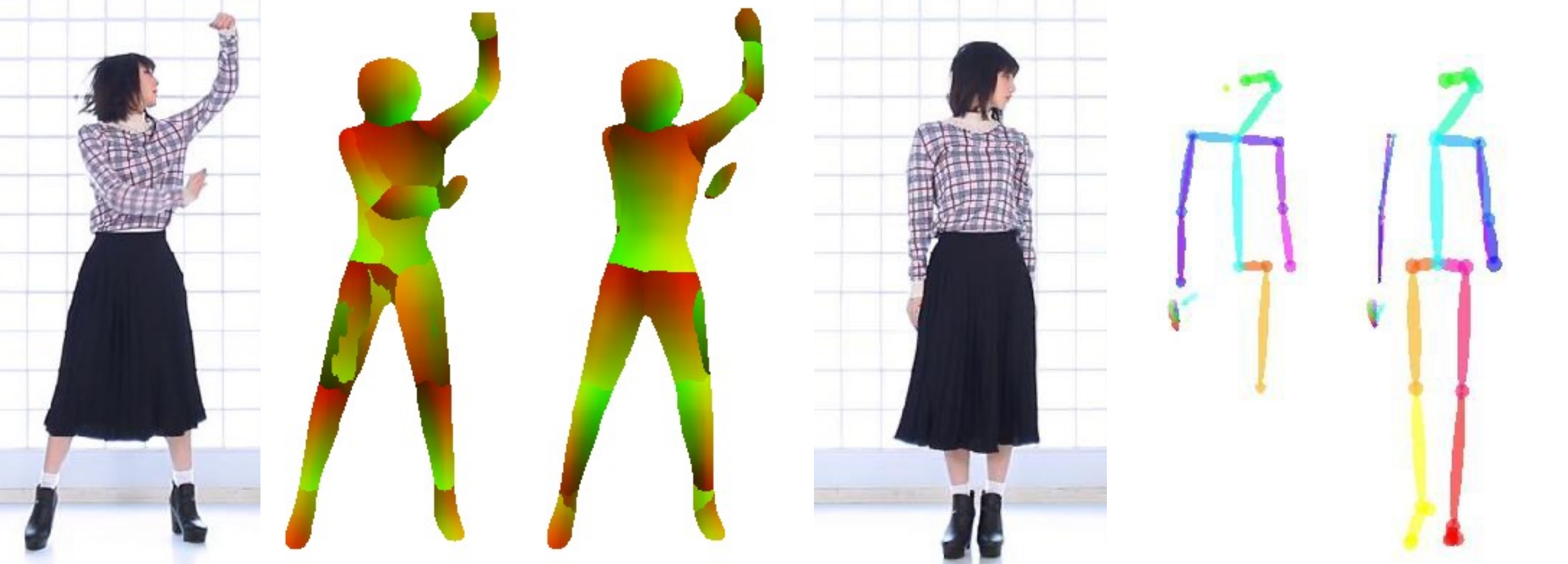}
  \caption{The current state of the art pose detection methods suffer from artifacts such as jitter, missing parts, and wrong detections. Given a reference image, we show the predictions for two consecutive frames. (Left) The dense body UV predicted by DensePose~\cite{densepose} has mistakes in the left arm and leg regions, it also misses the right arm. (Right) The keypoints predicted by OpenPose~\cite{openpose} miss the shoulder and the legs.}
  \label{fig:poseerror}
  \vspace*{-5mm}
\end{figure}

Our framework learns the dynamic appearance of a specific actor $\mathbb{X}$ from a reference video of the actor. We represent the reference video as a sequence of RGB image and pose pairs $\{\mathbb{I}_i, \mathbb{P}_i\}, i=1,\cdots N$. We assume the camera is fixed so the pose of the actor is represented by image-space pose representations such as 2D keypoints~\cite{openpose} and dense uv renderings~\cite{densepose}. At test time, given a motion sequence consisting of query poses $\{\mathbb{P}'_i\}, i=1,\cdots M$, the network synthesizes the corresponding appearance of $\mathbb{X}$ performing the motion. We adopt a StyleGAN based generator which has recently been shown to be effective in synthesizing high quality full body images~\cite{styleposegan}. In order to guide the generation of the actor performing a specific pose, we extract spatial pose features for each frame given $\mathbb{P}_i$ similar to~\cite{lewis2020vogue,styleposegan}. Unlike static image generation, however, the appearance of the garments, specifically if they are loose, are heavily affected by the motion of the actor. In order to capture such dynamic appearance changes (e.g., folds and wrinkles and secondary movement of loose clothes and hair), we introduce an explicit motion feature representation that is extracted from the past $K$ frames. We use such motion features as the latent style code to demodulate the generator. Furthermore, we observe two common limitations of the off-the-shelf pose estimators~\cite{openpose,densepose} as shown in Fig.~\ref{fig:poseerror}. First, being trained on single images, they often produce temporally jittery results when run on a video sequence. Second, in case of complex poses with self-occlusions, they are prone to missing parts and incorrect estimations. Since past frames provides strong cues to overcome these limitations, we propose a refinement module to regularize the spatial pose features based on the learned motion features. With the refined spatial pose features and the learned motion features, our generator synthesizes the final output image, $\mathbb{I}'_i$. Fig.~\ref{fig:arch} provides an overview of our pipeline.

\subsection{Input}
Given a sequence of RGB images $\{\mathbb{I}_i\}, i=1 \cdots N$, extracted from a captured video stream, we run DensePose \cite{densepose} to predict dense body IUV maps which are three channel images of the same size of the input image. We also use OpenPose \cite{openpose} to predict keypoints, i.e., skeleton, face, and hand landmarks, which are represented as RGB images of the same size $(W,H)$ as the input image. We concatenate the two sources of information to form a pose signature $\mathbb{P}_i \in \mathbb{R}^{6\times W \times H}$ for each input frame $i$. In order to extract explicit motion features, as explained in the next section, we also provide the pose signature of $K=10$ frames sampled unevenly from the past $20$ frames as input to our network. Specifically, we use the frames $\{1,2,3,4,6,8,10,13,16,20\}$ to build a motion signature $\mathbb{M}_i \in \mathbb{R}^{60 \times W \times H}$. The frames are sampled unevenly as closer frames have more impact on the current frame.

\subsection{Learning Pose/Motion Features}
Inspired by \cite{styleposegan}, we extract 2D spatial pose features $\mathcal{P}_i$ from the pose signature $\mathbb{P}_i$ to condition the generator to synthesize an image of the actor performing a specific pose. Since our generator is trained on a reference video of a specific actor, the global appearance of the actor is implicitly embedded into the generator weights. Unlike single-image generation case, however, the overall shape and appearance of the garments heavily depend on the motion being performed (e.g., the overall movement of a loose skirt, varying wrinkle and fold patterns). In order to capture such dynamic appearance changes, we propose to extract a 1D motion feature $\mathcal{M}_i$ from the motion signature $\mathbb{M}_i$ which are used to demodulate the generator via the AdaIN operations. To this end, we use a convolutional neural network $\mathbf{E}_\mathbf{P}$ to encode the pose signature $\mathbb{P}_i \in \mathbb{R}^{6\times W \times H}$ into a pose feature $\mathcal{P}_i \in \mathbb{R}^{512\times W_s \times H_s}$, where $W_s = W / 16, H_s = H/16$. Similarly, we extract the motion features with another convolutional neural network, $\mathbf{E}_\mathbf{M}$, which consists of a reshape operation and some fully connected layers to produce a 1D motion feature with a dimension of $2048$. We refer to the supplementary material for the detailed architecture.

\textbf{Temporal coherent refinement.} As shown in Fig.~\ref{fig:poseerror}, the current state of the art pose detection methods suffer from artifacts such as jitter, missing parts, and wrong detections. We hypothesize that observing the past motion, although imperfect, provides strong cues to identify and regularize such artifacts. Therefore, we propose to use the motion features $\mathcal{M}_i$ to refine the pose feature of the current frame $\mathcal{P}_i$. To realize such a refinement operation, we first concatenate $\mathcal{M}_i$ to each spatial location of $\mathcal{P}_i \in \mathbb{R}^{512\times W_s \times H_s}$ along the channel dimension to obtain an intermediate pose feature $\mathcal{P}_i^{int} \in \mathbb{R}^{2560\times W_s \times H_s}$. $\mathcal{P}_i^{int}$ is then passed through another convolutional neural network $\mathbf{E}_\mathbf{Refine}$ which produces a refined pose feature $\widetilde{\mathcal{P}_i} \in \mathbb{R}^{512\times W_s \times H_s}$. In Fig.~\ref{fig:xyplot} and Fig.~\ref{fig:flowmap}, we show that $\mathbf{E}_\mathbf{Refine}$ 
learns to effectively regularize the input pose of the current frame based on the motion feature which is learned from the past frames.

\subsection{Style-based Generator}
We use the style-based generator proposed by \cite{Karras20} for our image generation task. The original StyleGAN architecture takes a constant spatial tensor as input, and a latent noise vector passed through a mapping network to demodulate the intermediate layers to control the details of final generated image. As inspired by \cite{lewis2020vogue,styleposegan}, we use our learned spatial pose feature $\widetilde{\mathcal{P}_i}$ instead of the constant tensor to condition the generator. Different from previous work, we use the motion feature to control the dynamic appearance. Specifically, our generator, $\mathbf{T}(\widetilde{\mathcal{P}_i},\mathcal{M}_i)$, takes the spatial pose feature $\widetilde{\mathcal{P}_i}$ as input and passes it through a set of convolutional layers. The convolutional weights are then demodulated by the 1D motion feature $\mathcal{M}_i$. Our generator converts the spatial feature from size $512 \times W_s \times H_s$ to $3 \times W \times H$ after four residual blocks and four upsampling residual blocks, with random noise injection at every layer similar to~\cite{Karras20}. Therefore, our full pipeline can be written as:
\begin{equation}
    \mathbf{T}(\mathbf{E}_\mathbf{Refine}(\mathbf{E}_\mathbf{P}(\mathbb{P}_i),\mathbf{E}_\mathbf{M}(\mathbb{M}_i)) | \mathbf{E}_\mathbf{M}(\mathbb{M}_i)) = \mathbb{I}'_i
\end{equation}

\subsection{Loss Function}
We train our framework in an end-to-end manner. For a specific frame, $i$, in the training sequence, we directly supervise the generated image, $\mathbb{I}'_i$, with the ground truth image $\mathbb{I}_i$. Our loss function includes three terms. First, we use an $L1$ reconsturction loss:
\begin{equation}
    L_1 = |\mathbb{I}'_i - \mathbb{I}_i|.
\end{equation}

Second, we use the VGG-based perceptual loss~\cite{johnson2016perceptual} to encourage perceptual similarity:
\begin{equation}
    L_{VGG} = \sum_k {MSE}({VGG}_k(\mathbb{I}'_i) - {VGG}_k(\mathbb{I}_i)),
\end{equation}
where ${MSE(\cdot)}$ is the element-wise Mean-Square-Error, ${VGG_k}(\cdot)$ is the $k$-th layer of a VGG network pre-trained on ImageNet. Finally, we use an adversarial loss $L_{GAN}$ which utilizes a discriminator $\mathbf{D}$ identical to \cite{Karras20}.
During training, we minimize our objective $L = L_1 + L_{VGG} + L_{GAN}$ w.r.t. the parameters of $\mathbf{E}_\mathbf{P}$, $\mathbf{E}_\mathbf{M}$, $\mathbf{E}_\mathbf{Refine}$, and $\mathbf{T}$; and maximize $L_{GAN}$ w.r.t. $\mathbf{D}$.

\begin{figure*}[!h]
  \includegraphics[width=\textwidth]{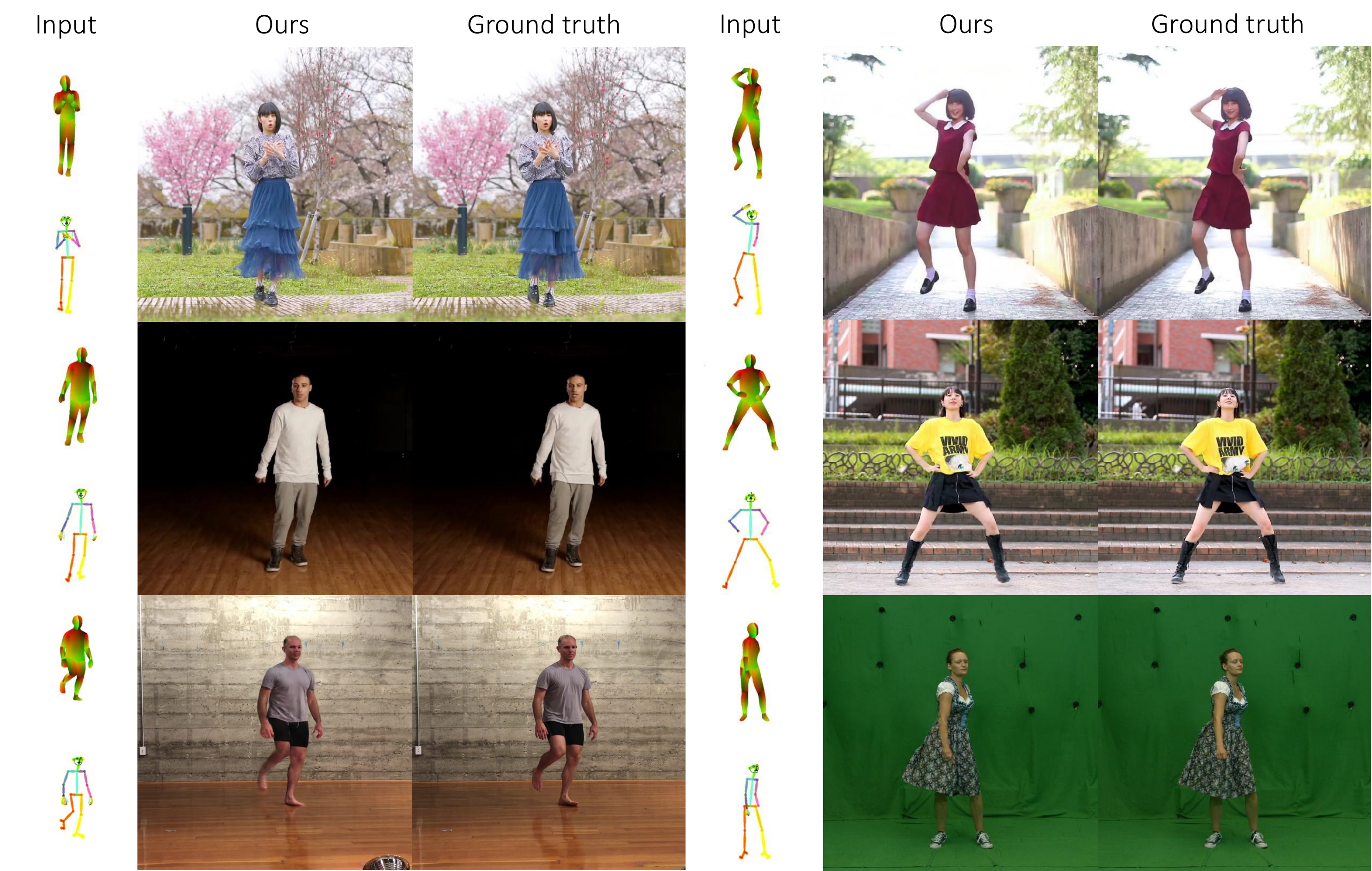}
  \caption{We train our network for character specific appearance synthesis. Here are the exemplary results on the test set for some sequences.}
  \label{fig:mainresults}
\end{figure*}

\section{Experiments}

We evaluate our approach on a set of Youtube videos with complex motion sequences as well as the datasets from existing work~\cite{chan2019dance,Kappel_2021_CVPR} (Sec.~\ref{sec:dataset}). We provide both quantitative and qualitative comparisons with previous work (Sec.~\ref{sec:evalandcomp}) and perform ablation studies to motivate the various design choices (Sec.~\ref{sec:ablation}).

We implement our network in Pytorch operating on images of resolution $512 \times 512$. We use the Adam optimizer with a learning rate of $0.02$. For a training sequence of $6K$ frames, the training takes approximately 72 hours for $100K$ iterations when trained with a batch size of $16$ on 4 NVIDIA V100 GPUS. Inference with our model takes
about 40 ms/frame on a single NVIDIA V100 GPU.

\subsection{Datasets}
\label{sec:dataset}

To test the performance of our method on unconstrained videos, we collect seven dancing sequences from Youtube. The selected videos are captured with a fixed camera, and show variation in terms of scenes (indoor vs. outdoor), garment types (tight clothing, loose or multilayer garments) and appearances (plain color, grid or stochastic texture, text logo). We also test our method on the sequences provided by \cite{chan2019dance} and \cite{Kappel_2021_CVPR}. In order to quantify the complexity of the motion sequences, we also calculate the corresponding motion speed.  Specifically, we calculate the average displacement of all keypoints between two consecutive frames which have been normalized to have a height of 1. In Table~\ref{Statistics} 
, we provide the different characteristics of each sequence along with the length of the sequences, i.e., the number of frames sampled with a frame rate of 24 fps. The speed of the motion for the sequences in our dataset are significantly higher than previous datasets demonstrating the complexity of our examples. The duration of a typical Youtube dancing video is between 2-5 minutes, which results in a sequence with 2k to 8k frames. This is significantly shorter than previous sequences captured in controlled lab settings, which often include more than 10k frames. We note that shorter sequences are more challenging test cases with respect to generalization to unseen poses.

\begin{table}[h!]
\resizebox{\columnwidth}{!}{
\begin{tabular}{c||c|c|c|c}
\hline
 & length & \makecell{motion \\ speed} & clothes type & texture \\ \hline \hline
Seq 1 & $7.5k$ & $3.7\times$ & loose & plain \\ \hline
Seq 2 & $3.4k$ & $2.9\times$ & loose & grid \\ \hline
Seq 3 & $6.0k$ & $3.0\times$ & loose & plain \\ \hline 
Seq 4 & $6.0k$ & $4.3\times$ & loose & stochastic \\ \hline
Seq 5 & $3.2k$ & $4.0\times$ & tight & text \\ \hline 
Seq 6 & $6.5k$ & $1.4\times$ & tight & plain \\ \hline 
Seq 7 & $6.1k$ & $4.5\times$ & multi-layer & stochastic \\ \hline
Seq 8 (\cite{Kappel_2021_CVPR}) & $12.5k$ & $0.006 (1.0\times)$ & loose & stochastic \\ \hline 
Seq 9 (\cite{chan2019dance}) & $11.4k$ & $2.7\times$ & tight & plain \\ \hline
\end{tabular}
}
\caption{Statistics of our dataset. For the motion speed of each sequence, we show the multiple of Seq 8.}
\label{Statistics}\small
\vspace*{-5mm}
\end{table}

\subsection{Evaluation and Comparisons}
\label{sec:evalandcomp}
For each sequence, we train our network with the first $85\%$ of the frames and test with the last $10\%$ of the frames. We skip the intermediate 5\% of frames to make sure that the beginning of the test sequence is different from the end of the train sequence.
In Fig~\ref{fig:mainresults}, we show example frames from the testing sequences where we synthesize the appearance of the character given different target poses and motion signatures. We refer to the supplementary video for more examples. 

\begin{figure*}[!h]

  \begin{overpic}[width=\textwidth]{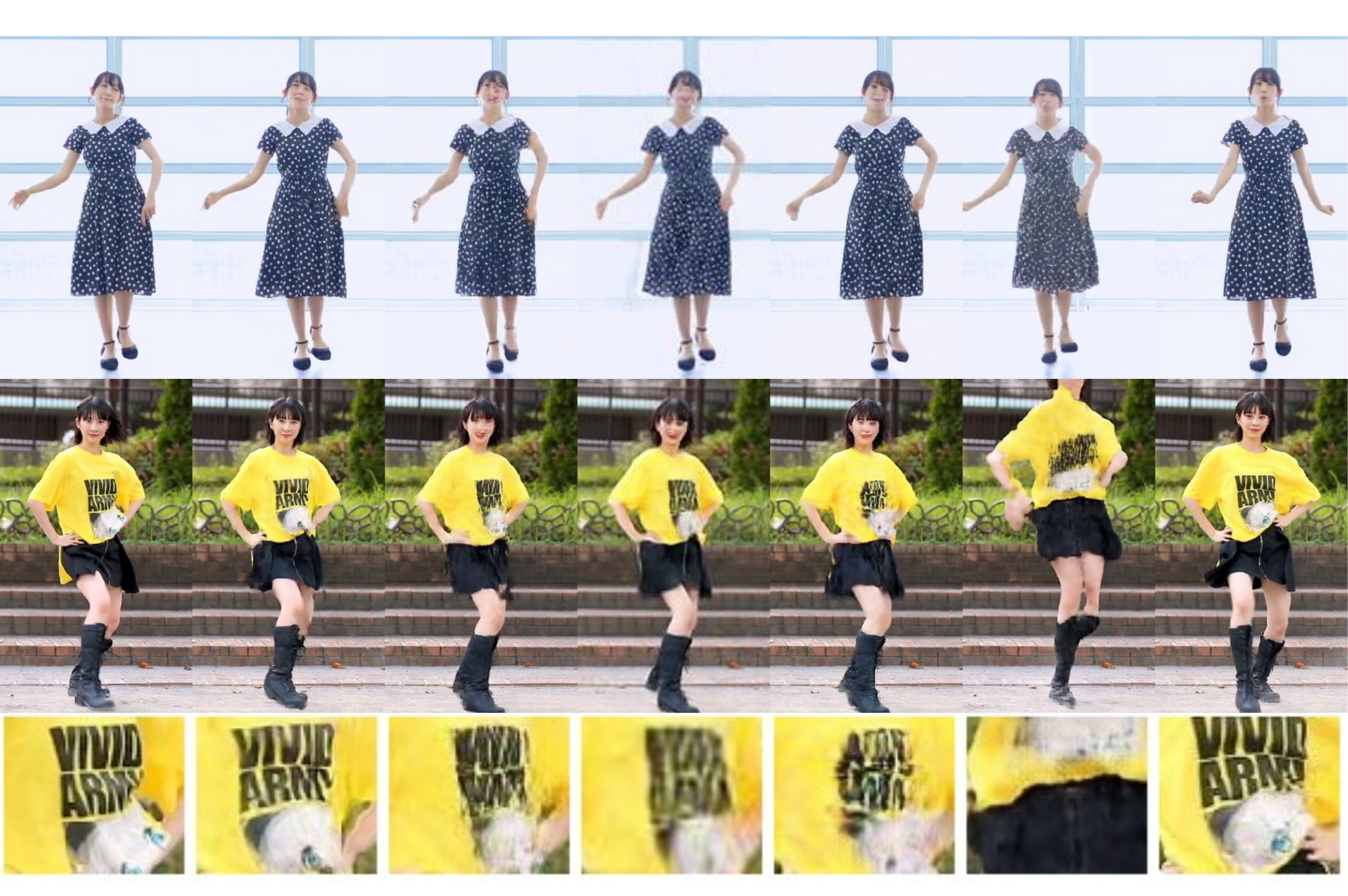}
    \put(2,65){Ground truth}
    \put(20,65){Ours}
    \put(30,65){pix2pixHD~\cite{pix2pix2017}}
    \put(46,65){vid2vid~\cite{wang2018video}}
    \put(62,65){EDN~\cite{chan2019dance}}
    \put(72,65){HF-NHMT~\cite{Kappel_2021_CVPR}}
    \put(86,65){Nearest neighbor}
  \end{overpic}

  \caption{We qualitatively compare our method with other related approaches. For the examples on the second row, we provide a close-up view for the text pattern on the shirt to highlight the synthesis quality.}
  \label{fig:baselines}
\end{figure*}
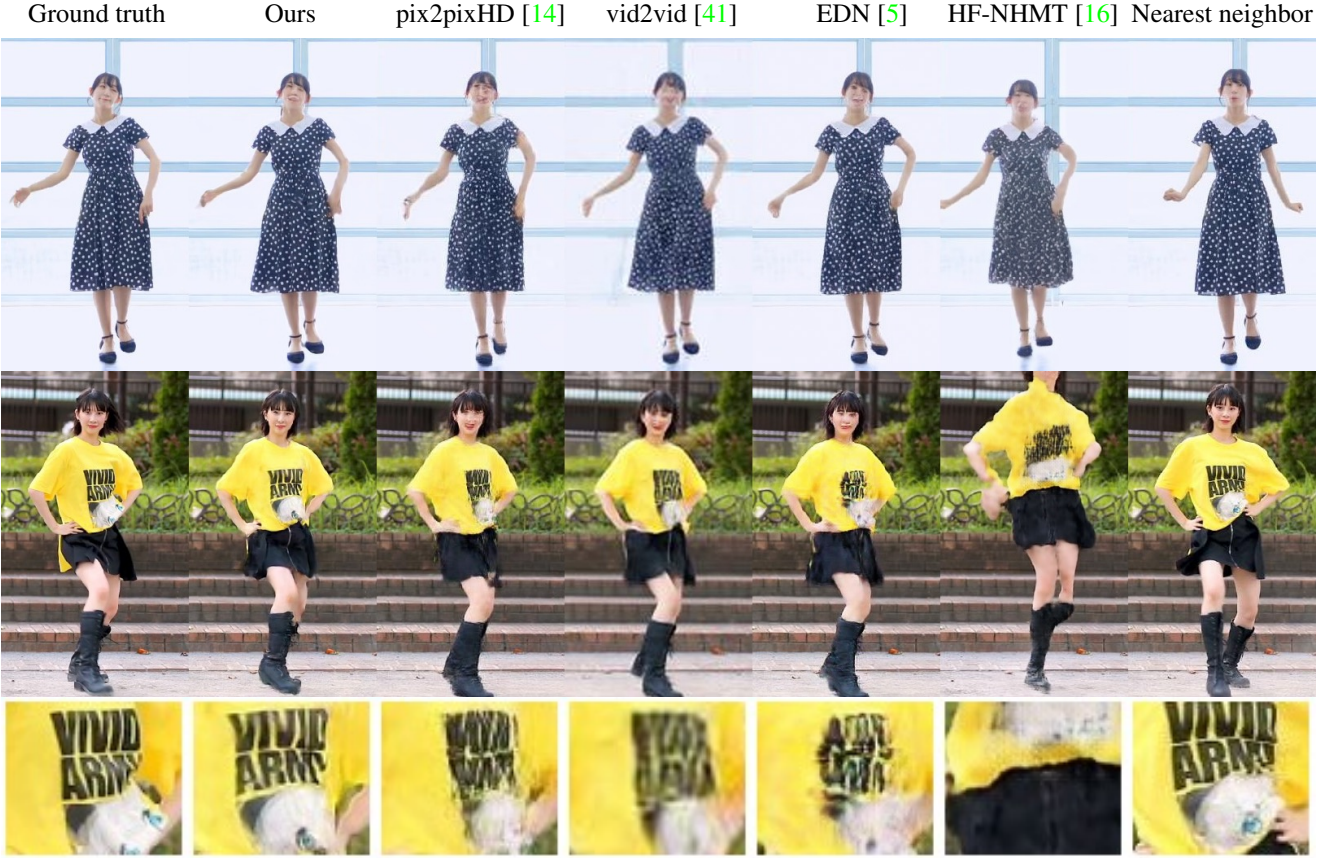

We also compare our approach with several previous approaches: (i) pix2pixHD~\cite{pix2pix2017}, the state-of-the-art image-to-image translation method that generates the full body appearance from dense uv based pose representations, (ii) its temporal extension vid2vid~\cite{wang2018video} which utilizes both keypoint and dense uv based posed representations, (iii) Everybody-Dance-Now (EDN)~\cite{chan2019dance} which takes 2D keypoints as input, and (iv) the recently proposed recurrent based architecture that focuses on actors with loose garments (HF-NHMT)~\cite{Kappel_2021_CVPR}. We also include the nearest neighbour (NN) baseline where for each query pose we select the frame from the training set with the most similar pose. We define the pose similarity as the L2 distance between the 2D keypoints. This baseline cannot capture the desired target poses since they are significantly different than the training poses. We train all the learning based alternatives on the same set of training frames as our method until convergence. In Fig~\ref{fig:baselines}, we provide visual results of our method, the baseline methods, and the corresponding ground truth. In Table~\ref{table:baseline}, we provide quantitative evaluations with respect to the ground truth using several metrics: (i) the mean square error (MSE) of the pixel value normalized to the range $[-1,1]$, (ii) the structural similarity index (SSIM)~\cite{wang2004image}, (iii) the perceptual similarity metric (LPIPS)~\cite{zhang2018unreasonable}, (iv) the Fréchet Inception Distance (FID)~\cite{heusel2017gans}, and (v) the tOF~\cite{chu2020learning}, pixel-wise difference of the estimated optical flow between each sequence and the ground truth. Our method outperforms the baselines with respect to all the metrics. Since our motion features $\mathcal{M}_i$ are learned from a short clip of past frames (a.k.a., motion window), it naturally encodes the temporal information. Therefore, we observe that our method generates more temporally smooth results without the need of an extra temporal discriminator. We also observe that conditioning the synthesis of the current frame on the motion features captures motion dependent appearance changes while avoiding the problem of error accumulation that is apparent in recurrent based approaches~\cite{Kappel_2021_CVPR}. 
\begin{table}[h!]
\resizebox{\columnwidth}{!}{
\begin{tabular}{c||c|c|c|c|c}
& MSE $\downarrow$ & SSIM $\uparrow$ & LPIPS $\downarrow$ & FID $\downarrow$ & tOF $\downarrow$\\
\hline \hline
pix2pixHD~\cite{pix2pix2017} & 0.0212 & 0.9807 & 0.0474 & 36.1323 & 7.6282\\
\hline 
vid2vid~\cite{wang2018video} & 0.0276 & 0.9795 & 0.2318 & 58.2085 & 5.5523\\
\hline
EDN~\cite{chan2019dance} & 0.0201 & 0.9811 & 0.0423 & 30.6969 & 6.1260\\
\hline
HF-NHMT~\cite{Kappel_2021_CVPR} & 0.0743 & 0.9629 & 0.1498 & 53.6912 & 8.5114\\
\hline
Ours &\textbf{0.0199} & \textbf{0.9813}& \textbf{0.0398}& \textbf{21.1877} & \textbf{5.4122}\\
\hline
\end{tabular}
}
\caption{Quantitative comparison to the related approaches. We evaluate the MSE, SSIM, LPIPS, FID, and tOF score of each method.}
\label{table:baseline}\small
\end{table}

\subsection{Ablation study}
\label{sec:ablation}
\begin{figure}[!t]
  \includegraphics[width=\columnwidth]{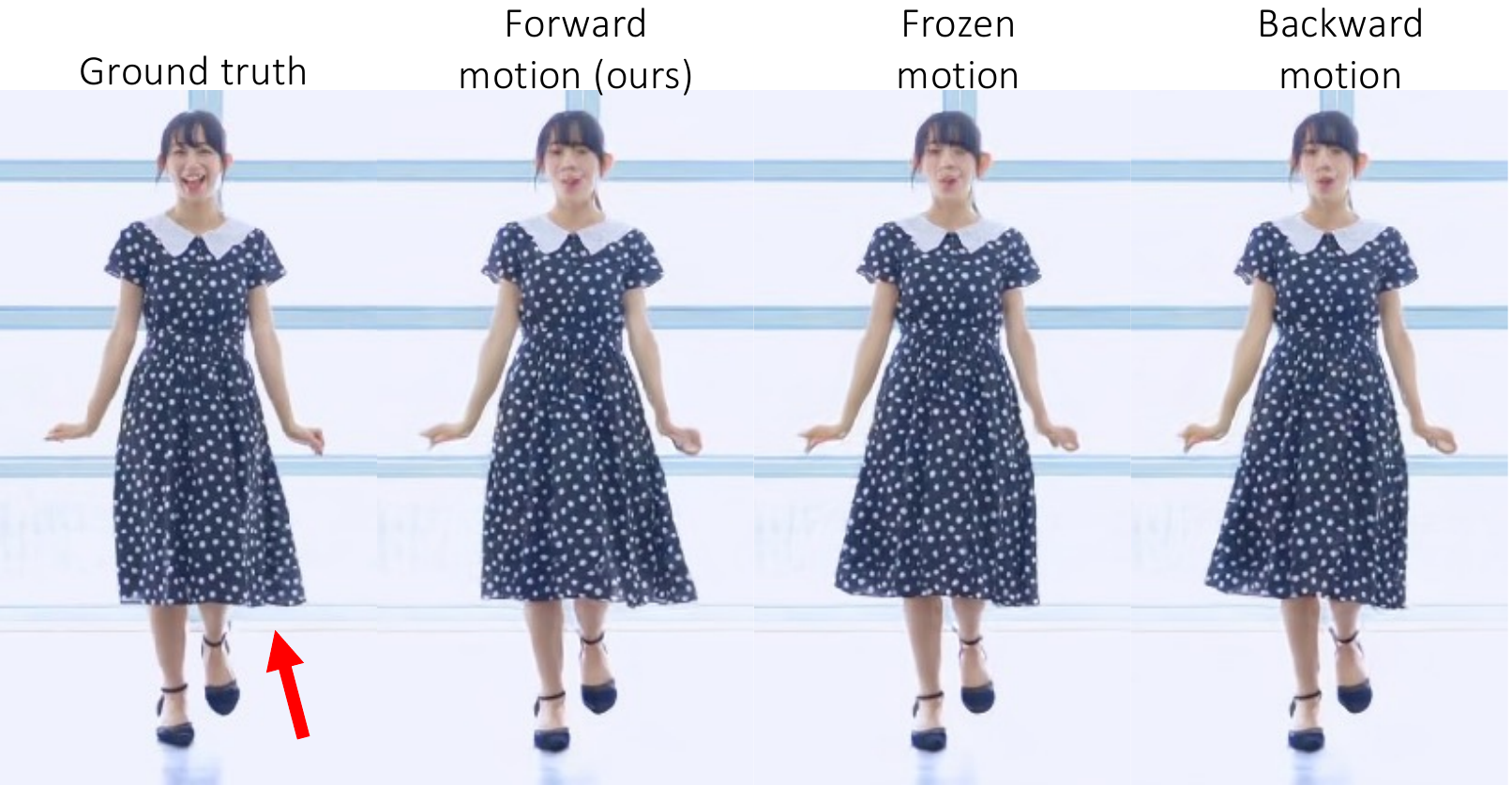}
  \caption{We show that our model captures motion specific appearance changes. We synthesize the same target frame by using a motion signature computed from (i) the original past frames, i.e., forward motion, (ii) from the same pose repeated for each past frame, i.e., frozen motion, and (iii) from the future frames in the reverse order, i.e., backward motion. In the forward motion, the character is swinging from her left side to right resulting in the skirt being dragged towards her left. While there is no significant dragging of the skirt with frozen motion signatures, the dragging is reversed with the backward motion.}
  \label{fig:disentangle}
 
\end{figure}

We perform various ablation studies to motivate our design choices. First, in order to demonstrate the effect of the motion features on capturing dynamic appearance changes, given the same pose signature $\mathbb{P}$, we synthesize frames with different motion signatures $\mathbb{M}$. Specifically, given a testing motion sequence, we synthesize a target frame by (i) computing the motion signature from the original sequence of past poses, i.e., \emph{forward} motion, (ii) hallucinate a still motion, i.e., a \emph{frozen} motion signature by using the same pose for all the past frames, and (iii) hallucinate the \emph{backward} motion, i.e., a motion signature computed from the future frames in the reverse order. As shown in Fig~\ref{fig:disentangle}, we show that our method is able to synthesize the overall shape of the garment that agree with the motion signature. While the flow of the skirt is reversed when simulating the backward motion, with frozen motion signatures, the skirt tends to keep a more stable rest shape.

\begin{figure}[!t]
  \includegraphics[width=\columnwidth]{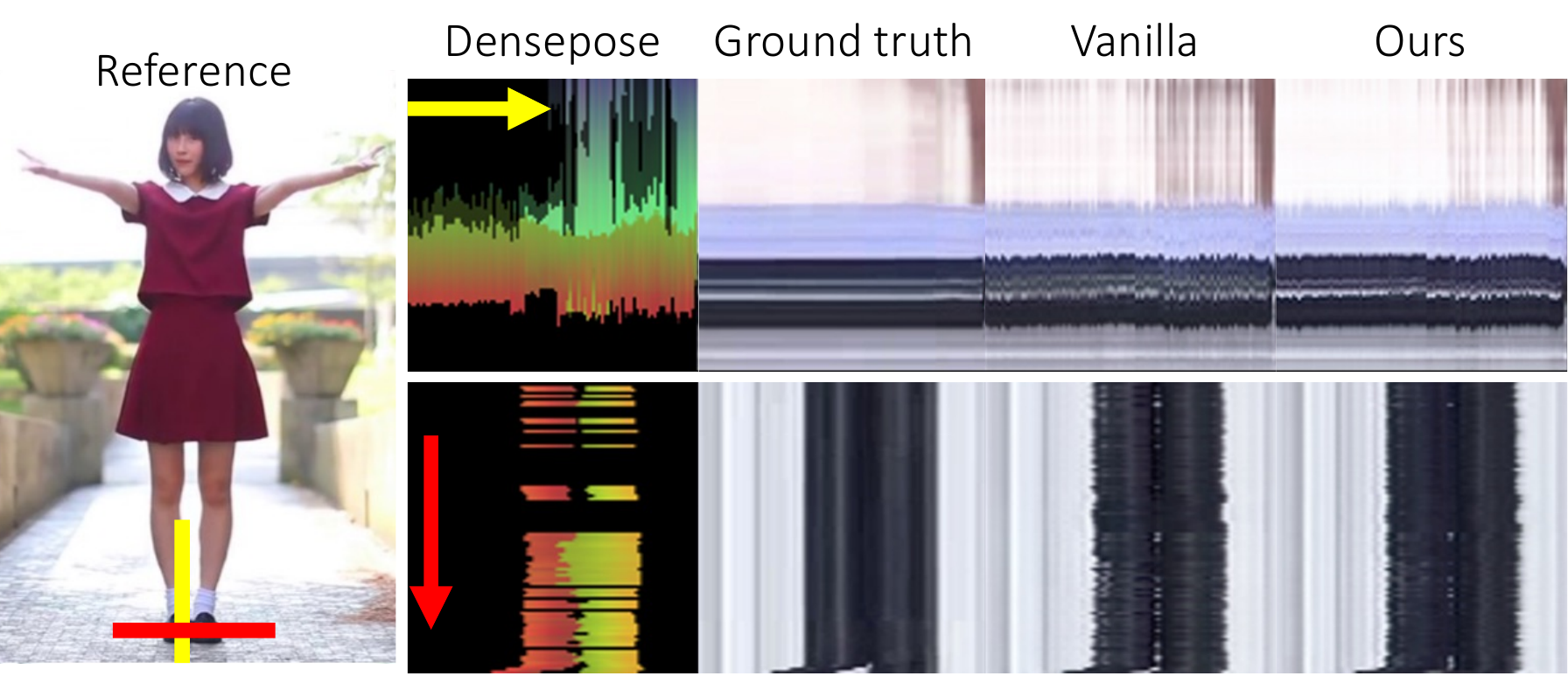}
  \caption{We cut a slice of vertical and horizontal pixels from a sequence of $100$ frames and concatenate them to form slice plots. Due to the artifacts in the dense pose estimation, the vanilla model generates jittery results. Ours refines the pose features and generates smoother results visually more similar to the ground truth.}
  \label{fig:xyplot}
   \vspace*{-5mm}
\end{figure}

Second, we study the effect of the length of the past frames used to compute the motion signature. Specifically, we compare four different cases where we use the (i) past $4$ frames, $\{1,2,3,4\}$; (ii) past $10$ frames, $\{1,2,3,4,6,8,10\}$; (iii) past $20$ frames, $\{1,2,3,4,6,8,10,13,16,20\}$ (the setting used in other experiments); and (iv) past $40$ frames, $\{1,2,3,4,6,8,10,13,16,20,24,29,34,40,47,56\}$. We calculate mean square error (MSE) and the structural similarity index (SSIM)~\cite{wang2004image} for the generated test sequences with respect to the ground truth. In Table~\ref{table:motionlength}, we show that using poses sampled in the past 20 frames provide a good balance. Shorter motion windows are not sufficient to capture motion dependent dynamic appearance changes. The trend shows the performance will decrease if the motion window gets shorter. When training without motion feature and pose refinement, the MSE and SSIM drops to 0.0213 and 0.9801 respectively. With longer motion windows, we observe no significant improvement in terms of capturing motion dependent appearance changes but training becomes more difficult due to the increase in the network size. Please also refer to the supplementary video for a qualitative comparison. 

\begin{table}[h!]
  \vspace*{2mm}
\resizebox{\columnwidth}{!}{
\centering
\begin{tabular}{c||c|c|c|c|c}
& 0 & 5 & 10 & 20 (ours) & 40\\
\hline \hline
MSE $\downarrow$ & 0.0213 & 0.0208 & 0.0201 & \textbf{0.0199} & 0.0205 \\
\hline 
SSIM $\uparrow$ & 0.9801 & 0.9809 & 0.9812 & \textbf{0.9813} & 0.9811\\
\hline
\end{tabular}
}
\vspace*{1mm}
\caption{ We evaluate the MSE and SSIM score for different length of the motion signature. Taking the past 20 frames is a good trade of between complexity and high-fidelity. }
\label{table:motionlength}\small
\end{table}

\begin{figure}[!t]
  \includegraphics[width=\columnwidth]{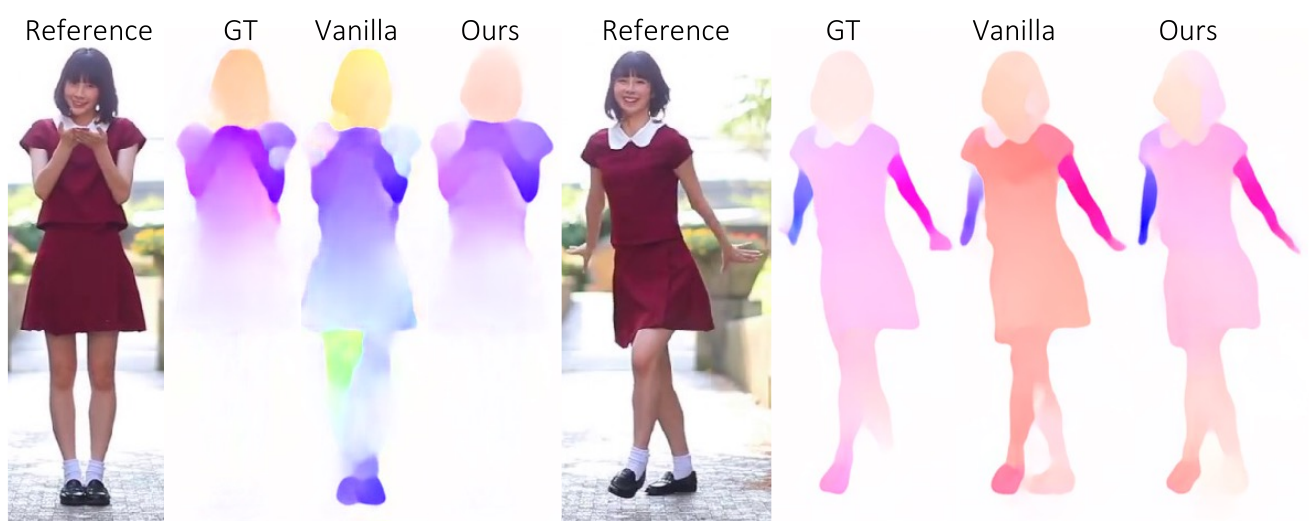}
  \caption{We use~\cite{ilg2017flownet} to compute the flow map between the consecutive frames in the ground truth and generated sequences with ours and the vanilla model. Our approach better captures the motion of the ground truth sequence.}
  \label{fig:flowmap}
  \vspace*{-5mm}
\end{figure}

\begin{figure*}[!h]
  \includegraphics[width=\textwidth]{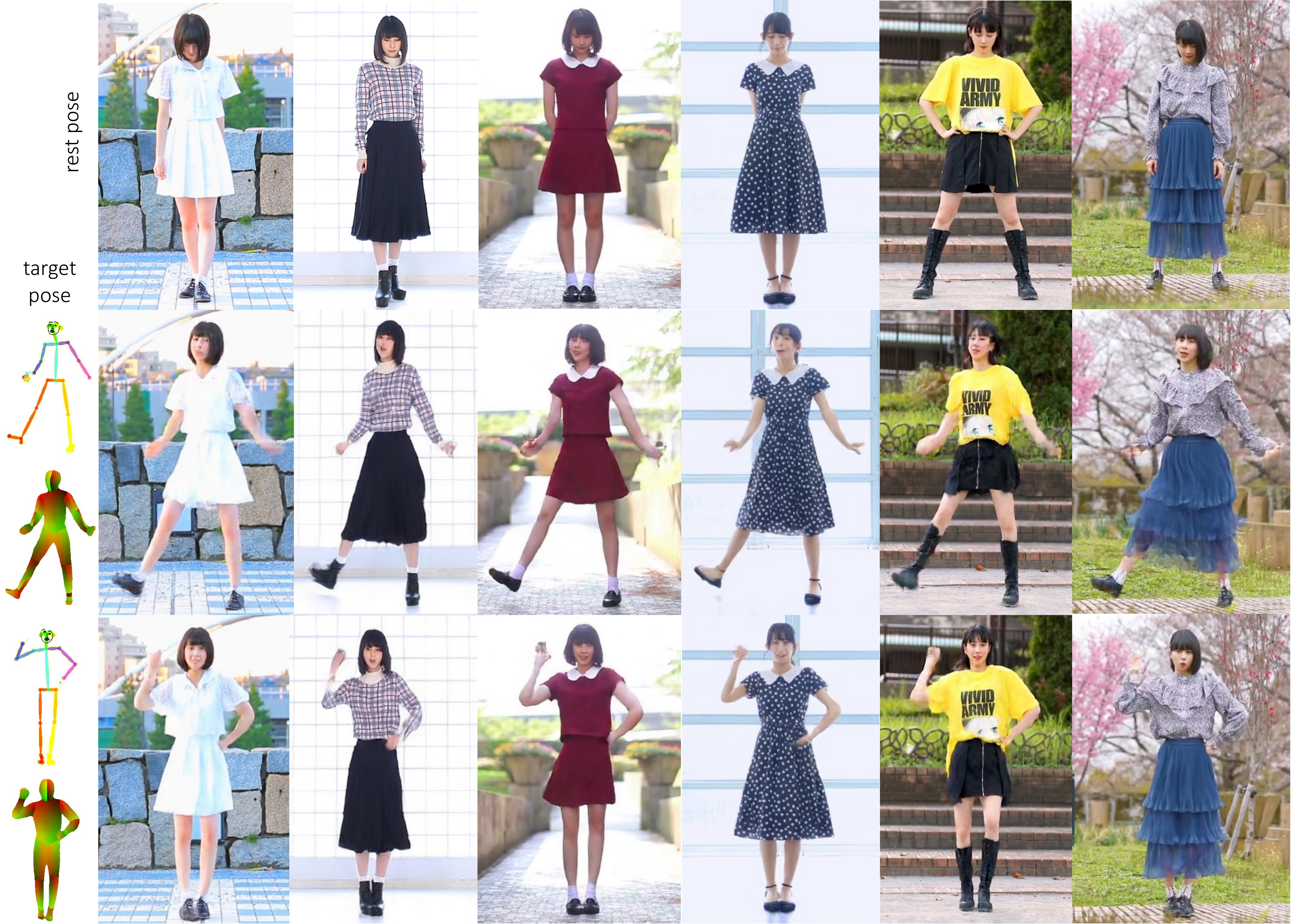}
  \caption{Our approach allows motion transfer from a source sequence to a target actor.}
  \label{fig:transfer}
\end{figure*}

Finally, we study the effect of our motion driven pose refinement on temporal coherency. Specifically, we compare our proposed network with a vanilla motion driven pipeline which directly feeds the generator $\mathbf{T}$ with the learned pose feature $\mathcal{P}_i$, instead of the refined features $\widetilde{\mathcal{P}_i}$. We find that the vanilla motion driven approach achieves similar per-frame image synthesis quality but suffers from significant temporal coherency artifacts. In Fig.~\ref{fig:xyplot}, we identify a vertical and a horizontal slice of pixels in a particular frame and concatenate such slices over time (for $100$ frames) along the dimension denoted by the arrows to form a slice plot. The corresponding video clip does not contain significant movements around the leg resulting in smooth ground truth slice plots along the time axis. However, the dense body UV predictions~\cite{densepose} are not temporally stable producing high frequency signals in the corresponding densepose slice plots. The vanilla motion driven pipeline is affected by such artifacts and reproduces the high frequency signals, while ours with pose refinement produces a significantly smoother slice plot. We also present the predicted flow map~\cite{ilg2017flownet} between the consecutive frames in the ground truth and generated sequences in Fig.~\ref{fig:flowmap} to show that our approach best captures the motion of the ground truth regardless the imperfect pose input. The average mean square error between the ground truth and generated flow maps also show the advantage of our method, ours and the vanilla version achieve an error of $5.41$ and $6.11$ respectively.

Please refer to our supplementary material for more evaluation and experiments.
\section{Application: Motion transfer}
Once we have an actor specific network, a typical application is to retarget a source motion extracted from a video of another actor. Since the two actors might have different body proportions, before retargeting, we first perform a simple alignment. We adjust the height and width of the detected skeletons and move them vertically so that they stand on the ground of the target background image~\cite{chan2019dance}. In Fig.~\ref{fig:transfer}, we show some examples, please refer to our supplementary video for more results.

\section{Conclusion, Limitations and Future Work}
  \vspace*{-2mm}
In this paper, we present a StyleGAN based approach for video synthesis of an actor performing complex motion sequences. We propose a novel motion signature which is used to modulate the generator to capture dynamic appearance changes as well as refining the pose features to obtain temporally coherent results. Our approach outperforms the previous alternatives both quantitatively and qualitatively.

While showing impressive results, our method has some limitations. We use image based pose representations which are limited to capture poses such as 3D rotations. Incorporating 3D representations, e.g., SMPL~\cite{loper2015smpl}, is an interesting direction. The quality of the results can degrade with very complex motions significantly different than training sequences (see the supplementary material). Improving motion generalization is a promising direction. We want to extend our work to videos captured with moving cameras and incorporate the motion of the camera into the generation process. We present a simple retargeting and alignment process for motion transfer. More sophisticated retargeting strategies are needed to transfer motion between actors with significantly different body shapes. While our method generates reasonable face and hand regions, further refining these body parts with specialized modules is worth investigating. Currently the model is trained for a specific identity, supporting multiple identities by conditioning our network on an extra channel that denotes the character identity is an interesting direction for future work. Finally, adopting advanced GAN architectures, e.g., Alias-Free GAN~\cite{Karras2021}, may further improve the performance without much change in the overall design. 

\textbf{Acknowledgement.} We would like to thank the anonymous reviewers for their constructive comments; as well as Jae Shin Yoo, Moritz Kappel, and Erika Lu for their kind help with the experiments and discussions.

{\small
\bibliographystyle{ieee_fullname}
\bibliography{egpaper_final}

\begin{thebibliography}{10}\itemsep=-1pt

\bibitem{Aberman2019}
K. Aberman, M. Shi, J. Liao, D. Lischinski, B. Chen, and D. Cohen-Or.
\newblock Deep video-based performance cloning.
\newblock {\em Computer Graphics Forum}, 38(2):219--233, 2019.

\bibitem{RecycleGAN}
Aayush Bansal, Shugao Ma, Deva Ramanan, and Yaser Sheikh.
\newblock Recycle-gan: Unsupervised video retargeting.
\newblock In {\em ECCV}, 2018.

\bibitem{openpose}
Z. {Cao}, G. {Hidalgo Martinez}, T. {Simon}, S. {Wei}, and Y.~A. {Sheikh}.
\newblock Openpose: Realtime multi-person 2d pose estimation using part
  affinity fields.
\newblock {\em IEEE Transactions on Pattern Analysis and Machine Intelligence},
  2019.

\bibitem{Casas2014}
Dan Casas, Marco Volino, John Collomosse, and Adrian Hilton.
\newblock 4d video textures for interactive character appearance.
\newblock {\em CGF}, 33(2):371–380, May 2014.

\bibitem{chan2019dance}
Caroline Chan, Shiry Ginosar, Tinghui Zhou, and Alexei~A Efros.
\newblock Everybody dance now.
\newblock In {\em ICCV}, 2019.

\bibitem{chu2020learning}
Mengyu Chu, You Xie, Jonas Mayer, Laura Leal-Taix{\'e}, and Nils Thuerey.
\newblock Learning temporal coherence via self-supervision for gan-based video
  generation.
\newblock {\em ACM Transactions on Graphics (TOG)}, 39(4):75--1, 2020.

\bibitem{Collet:2015}
Alvaro Collet, Ming Chuang, Pat Sweeney, Don Gillett, Dennis Evseev, David
  Calabrese, Hugues Hoppe, Adam Kirk, and Steve Sullivan.
\newblock High-quality streamable free-viewpoint video.
\newblock {\em ACM Trans. Graph.}, 34(4), July 2015.

\bibitem{Grigorev_2021_CVPR}
Artur Grigorev, Karim Iskakov, Anastasia Ianina, Renat Bashirov, Ilya
  Zakharkin, Alexander Vakhitov, and Victor Lempitsky.
\newblock Stylepeople: A generative model of fullbody human avatars.
\newblock In {\em Proceedings of the IEEE/CVF Conference on Computer Vision and
  Pattern Recognition (CVPR)}, pages 5151--5160, June 2021.

\bibitem{grigorev2019coordinate}
Artur Grigorev, Artem Sevastopolsky, Alexander Vakhitov, and Victor Lempitsky.
\newblock Coordinate-based texture inpainting for pose-guided human image
  generation.
\newblock In {\em Proceedings of the IEEE/CVF Conference on Computer Vision and
  Pattern Recognition}, pages 12135--12144, 2019.

\bibitem{densepose}
Riza~Alp G\"uler, Natalia Neverova, and Iasonas Kokkinos.
\newblock Densepose: Dense human pose estimation in the wild.
\newblock 2018.

\bibitem{habermann2021}
Marc Habermann, Lingjie Liu, Weipeng Xu, Michael Zollhoefer, Gerard Pons-Moll,
  and Christian Theobalt.
\newblock Real-time deep dynamic characters.
\newblock {\em ACM TOG}, 40(4), aug 2021.

\bibitem{heusel2017gans}
Martin Heusel, Hubert Ramsauer, Thomas Unterthiner, Bernhard Nessler, and Sepp
  Hochreiter.
\newblock Gans trained by a two time-scale update rule converge to a local nash
  equilibrium.
\newblock {\em Advances in neural information processing systems}, 30, 2017.

\bibitem{ilg2017flownet}
Eddy Ilg, Nikolaus Mayer, Tonmoy Saikia, Margret Keuper, Alexey Dosovitskiy,
  and Thomas Brox.
\newblock Flownet 2.0: Evolution of optical flow estimation with deep networks.
\newblock In {\em Proceedings of the IEEE conference on computer vision and
  pattern recognition}, pages 2462--2470, 2017.

\bibitem{pix2pix2017}
Phillip Isola, Jun-Yan Zhu, Tinghui Zhou, and Alexei~A Efros.
\newblock Image-to-image translation with conditional adversarial networks.
\newblock {\em CVPR}, 2017.

\bibitem{johnson2016perceptual}
Justin Johnson, Alexandre Alahi, and Li Fei-Fei.
\newblock Perceptual losses for real-time style transfer and super-resolution.
\newblock In {\em European conference on computer vision}, pages 694--711.
  Springer, 2016.

\bibitem{Kappel_2021_CVPR}
Moritz Kappel, Vladislav Golyanik, Mohamed Elgharib, Jann-Ole Henningson,
  Hans-Peter Seidel, Susana Castillo, Christian Theobalt, and Marcus Magnor.
\newblock High-fidelity neural human motion transfer from monocular video.
\newblock In {\em Proceedings of the IEEE/CVF Conference on Computer Vision and
  Pattern Recognition (CVPR)}, pages 1541--1550, June 2021.

\bibitem{Karras2021}
Tero Karras, Miika Aittala, Samuli Laine, Erik H\"{a}rk\"{o}nen, Janne
  Hellsten, Jaakko Lehtinen, and Timo Aila.
\newblock Alias-free generative adversarial networks.
\newblock {\em CoRR}, abs/2106.12423, 2021.

\bibitem{Karras19}
Tero Karras, Samuli Laine, and Timo Aila.
\newblock A style-based generator architecture for generative adversarial
  networks.
\newblock In {\em CVPR}, pages 4401--4410. Computer Vision Foundation / {IEEE},
  2019.

\bibitem{Karras20}
Tero Karras, Samuli Laine, Miika Aittala, Janne Hellsten, Jaakko Lehtinen, and
  Timo Aila.
\newblock Analyzing and improving the image quality of stylegan.
\newblock In {\em CVPR}, volume abs/1912.04958, 2019.

\bibitem{lewis2021tryongan}
Kathleen~M Lewis, Srivatsan Varadharajan, and Ira Kemelmacher-Shlizerman.
\newblock Tryongan: Body-aware try-on via layered interpolation.
\newblock {\em ACM SIGGRAPH}, 40(4), 2021.

\bibitem{lewis2020vogue}
Kathleen~M Lewis, Srivatsan Varadharajan, and Ira Kemelmacher-Shlizerman.
\newblock Vogue: Try-on by stylegan interpolation optimization.
\newblock {\em arXiv preprint arXiv:2101.02285}, 2021.

\bibitem{liu2020NeuralHumanRendering}
Lingjie Liu, Weipeng Xu, Marc Habermann, Michael Zollhöfer, Florian Bernard,
  Hyeongwoo Kim, Wenping Wang, and Christian Theobalt.
\newblock Neural human video rendering by learning dynamic textures and
  rendering-to-video translation.
\newblock {\em IEEE TVCG}, PP:1--1, 05 2020.

\bibitem{Liu2019Neural}
Lingjie Liu, Weipeng Xu, Michael Zollhoefer, Hyeongwoo Kim, Florian Bernard,
  Marc Habermann, Wenping Wang, and Christian Theobalt.
\newblock Neural rendering and reenactment of human actor videos.
\newblock {\em ACM TOG}, 2019.

\bibitem{lwb2019}
Wen Liu, Zhixin Piao, Min Jie, Wenhan Luo, Lin Ma, and Shenghua Gao.
\newblock Liquid warping gan: A unified framework for human motion imitation,
  appearance transfer and novel view synthesis.
\newblock In {\em The IEEE International Conference on Computer Vision (ICCV)},
  2019.

\bibitem{loper2015smpl}
Matthew Loper, Naureen Mahmood, Javier Romero, Gerard Pons-Moll, and Michael~J
  Black.
\newblock Smpl: A skinned multi-person linear model.
\newblock {\em ACM transactions on graphics (TOG)}, 34(6):1--16, 2015.

\bibitem{ma2017pose}
Liqian Ma, Xu Jia, Qianru Sun, Bernt Schiele, Tinne Tuytelaars, and Luc
  Van~Gool.
\newblock Pose guided person image generation.
\newblock In {\em NeurIPS}, pages 405--415, 2017.

\bibitem{ma2017disentangled}
Liqian Ma, Qianru Sun, Stamatios Georgoulis, Luc Van~Gool, Bernt Schiele, and
  Mario Fritz.
\newblock Disentangled person image generation.
\newblock In {\em CVPR}, June 2018.

\bibitem{neverova2018}
Natalia Neverova, Riza~Alp G{\"u}ler, and Iasonas Kokkinos.
\newblock {Dense Pose Transfer}.
\newblock In {\em ECCV}, Munich, Germany, Sept. 2018.

\bibitem{peng2021neural}
Sida Peng, Yuanqing Zhang, Yinghao Xu, Qianqian Wang, Qing Shuai, Hujun Bao,
  and Xiaowei Zhou.
\newblock Neural body: Implicit neural representations with structured latent
  codes for novel view synthesis of dynamic humans.
\newblock In {\em CVPR}, 2021.

\bibitem{SMPLpix:WACV:2020}
Sergey Prokudin, Michael~J. Black, and Javier Romero.
\newblock {SMPLpix}: Neural avatars from {3D} human models.
\newblock In {\em Winter Conference on Applications of Computer Vision (WACV)},
  pages 1810--1819, Jan. 2021.

\bibitem{PumarolaASM18}
Albert Pumarola, Antonio Agudo, Alberto Sanfeliu, and Francesc Moreno{-}Noguer.
\newblock Unsupervised person image synthesis in arbitrary poses.
\newblock In {\em CVPR}, pages 8620--8628. {IEEE} Computer Society, 2018.

\bibitem{raj2021anr}
Amit Raj, Julian Tanke, James Hays, Minh Vo, Carsten Stoll, and Christoph
  Lassner.
\newblock Anr-articulated neural rendering for virtual avatars.
\newblock In {\em arXiv:2012.12890}, 2021.

\bibitem{styleposegan}
Kripasindhu Sarkar, Vladislav Golyanik, Lingjie Liu, and Christian Theobalt.
\newblock Style and pose control for image synthesis of humans from a single
  monocular view, 2021.

\bibitem{Sarkar2020}
Kripasindhu Sarkar, Dushyant Mehta, Weipeng Xu, Vladislav Golyanik, and
  Christian Theobalt.
\newblock Neural re-rendering of humans from a single image.
\newblock In {\em European Conference on Computer Vision (ECCV)}, 2020.

\bibitem{Shysheya_2019_CVPR}
Aliaksandra Shysheya, Egor Zakharov, Kara-Ali Aliev, Renat Bashirov, Egor
  Burkov, Karim Iskakov, Aleksei Ivakhnenko, Yury Malkov, Igor Pasechnik,
  Dmitry Ulyanov, Alexander Vakhitov, and Victor Lempitsky.
\newblock Textured neural avatars.
\newblock In {\em CVPR}, June 2019.

\bibitem{Siarohin_2018_CVPR}
Aliaksandr Siarohin, Enver Sangineto, Stéphane Lathuilière, and Nicu Sebe.
\newblock Deformable gans for pose-based human image generation.
\newblock In {\em CVPR)}, June 2018.

\bibitem{tewari2020stylerig}
Ayush Tewari, Mohamed Elgharib, Gaurav Bharaj, Florian Bernard, Hans-Peter
  Seidel, Patrick P{\'e}rez, Michael Z{\"o}llhofer, and Christian Theobalt.
\newblock Stylerig: Rigging stylegan for 3d control over portrait images, cvpr
  2020.
\newblock In {\em {IEEE} Conference on Computer Vision and Pattern Recognition
  (CVPR)}. {IEEE}, june 2020.

\bibitem{Tewari2020}
A. Tewari, O. Fried, J. Thies, V. Sitzmann, S. Lombardi, K. Sunkavalli, R.
  Martin-Brualla, T. Simon, J. Saragih, M. Nießner, R. Pandey, S. Fanello, G.
  Wetzstein, J.-Y. Zhu, C. Theobalt, M. Agrawala, E. Shechtman, D.~B Goldman,
  and M. Zollhöfer.
\newblock State of the art on neural rendering.
\newblock {\em CGF}, 39(2):701--727, 2020.

\bibitem{thies2019deferred}
Justus Thies, Michael Zollh{\"o}fer, and Matthias Nie{\ss}ner.
\newblock Deferred neural rendering: Image synthesis using neural textures.
\newblock {\em ACM TOG}, 38(4):1--12, 2019.

\bibitem{Volino:BMVC:2014}
M. Volino, D. Casas, J.P. Collomosse, and A. Hilton.
\newblock Optimal representation of multiple view video.
\newblock 2014.

\bibitem{wang2018video}
Ting-Chun Wang, Ming-Yu Liu, Jun-Yan Zhu, Guilin Liu, Andrew Tao, Jan Kautz,
  and Bryan Catanzaro.
\newblock Video-to-video synthesis.
\newblock {\em arXiv preprint arXiv:1808.06601}, 2018.

\bibitem{wang2004image}
Zhou Wang, Alan~C Bovik, Hamid~R Sheikh, and Eero~P Simoncelli.
\newblock Image quality assessment: from error visibility to structural
  similarity.
\newblock {\em IEEE transactions on image processing}, 13(4):600--612, 2004.

\bibitem{zhang2021dynamic}
Meng Zhang, Duygu Ceylan, Tuanfeng Wang, and Niloy~J Mitra.
\newblock Dynamic neural garments.
\newblock {\em arXiv preprint arXiv:2102.11811}, 2021.

\bibitem{zhang2018unreasonable}
Richard Zhang, Phillip Isola, Alexei~A Efros, Eli Shechtman, and Oliver Wang.
\newblock The unreasonable effectiveness of deep features as a perceptual
  metric.
\newblock In {\em Proceedings of the IEEE conference on computer vision and
  pattern recognition}, pages 586--595, 2018.

\bibitem{zhu2019progressive}
Zhen Zhu, Tengteng Huang, Baoguang Shi, Miao Yu, Bofei Wang, and Xiang Bai.
\newblock Progressive pose attention transfer for person image generation.
\newblock In {\em CVPR}, pages 2347--2356, 2019.

\end{thebibliography}
}

\end{document}